\documentclass[11pt,a4paper]{article}
\usepackage[hyperref]{emnlp2020}
\usepackage{times}
\usepackage{latexsym}
\usepackage{amsmath}
\usepackage{pifont}

\usepackage{graphicx}
\usepackage{booktabs}
\usepackage{multirow}
\usepackage{tikz}
\usepackage{amssymb}
\usepackage[T1]{fontenc}
\usepackage{textcomp}

\newcommand{\cmark}{\ding{51}}
\newcommand{\xmark}{\ding{55}}

\usepackage{microtype}
\usepackage{xcolor}
\usepackage{soul}
\newcommand{\mathcolorbox}[2]{\colorbox{#1}{$\displaystyle #2$}}

\colorlet{soulred}{red!30}
\DeclareRobustCommand{\hlred}[1]{{\sethlcolor{soulred}\hl{#1}}}

\colorlet{soulbleu}{cyan!20}
\DeclareRobustCommand{\hlblue}[1]{{\sethlcolor{soulbleu}\hl{#1}}}

\colorlet{soulgreen}{green!20}
\DeclareRobustCommand{\hlgreen}[1]{{\sethlcolor{soulgreen}\hl{#1}}}

\colorlet{soulyellow}{yellow!40}
\DeclareRobustCommand{\hlyellow}[1]{{\sethlcolor{soulyellow}\hl{#1}}}

\colorlet{soulorange}{orange!30}
\DeclareRobustCommand{\hlorange}[1]{{\sethlcolor{soulorange}\hl{#1}}}

\colorlet{soulpurple}{blue!30}

\aclfinalcopy 

\title{MinTL: Minimalist Transfer Learning for Task-Oriented \\Dialogue Systems}

\author{Zhaojiang Lin, Andrea Madotto, Genta Indra Winata, Pascale Fung \\
Center for Artificial Intelligence Research (CAiRE)\\
  Department of Electronic and Computer Engineering\\
  The Hong Kong University of Science and Technology, Clear Water Bay, Hong Kong\\
\texttt{zlinao@connect.ust.hk}}

\date{}

\begin{document}
\maketitle

\begin{abstract}
In this paper, we propose Minimalist Transfer Learning (\textit{MinTL}) to simplify the system design process of task-oriented dialogue systems and alleviate the over-dependency on annotated data. \textit{MinTL} is a simple yet effective transfer learning framework, which allows us to plug-and-play pre-trained seq2seq models, and jointly learn dialogue state tracking and dialogue response generation. 
Unlike previous approaches, which use a copy mechanism to ``carryover'' the old dialogue states to the new one, we introduce \textit{Levenshtein belief spans} ($Lev$), that allows efficient dialogue state tracking with a minimal generation length. We instantiate our learning framework with two pre-trained backbones: T5~\cite{raffel2019exploring} and BART~\cite{lewis2019bart}, and evaluate them on MultiWOZ. Extensive experiments demonstrate that: 1) our systems establish new state-of-the-art results on end-to-end response generation, 2) \textit{MinTL}-based systems are more robust than baseline methods in the low resource setting, and they achieve competitive results with only 20\% training data, and 3) $Lev$ greatly improves the inference efficiency\footnote{Code available in \url{https://github.com/zlinao/MinTL}}.
\end{abstract}

\section{Introduction}
Building robust task-oriented dialogue systems is challenging due to complex system design and limited availability of human-annotated data~\cite{wen2017network,wu2019alternating}. A dialogue agent is expected to learn dialogue reasoning, decision making, and language generation, which require a large amount of training data. 
However, collecting and annotating data for training a dialogue system is time-intensive and not transferable among domains~\cite{young2013pomdp}. One possible workaround is to leverage the pre-trained language model to reduce human supervision~\cite{budzianowski2019hello}. 

Recent progress in pre-training language models has been shown to be promising in alleviating the data scarcity problem~\cite{budzianowski2019hello,wu2020tod}.
 Such models are typically pre-trained on large-scale plain text with self-supervised objectives, e.g., language modeling~\cite{radford2019language} and language denoising~\cite{devlin2019bert}. Fine tuning pre-trained language models improves a wide range of natural language processing applications~\cite{lewis2019bart,raffel2019exploring}, notably machine translation~\cite{conneau2019cross}, and personalized dialogue response generation~\cite{wolf2019transfertransfo}. However, adapting pre-trained language models to task-oriented dialogue systems is not trivial. 
 Current state-of-the-art (SOTA) approaches in task-oriented dialogue rely on several tasks-specific modules, such as State Operation Predictor~\cite{kim2019efficient} for dialogue state tracking, and CopyNet~\cite{gu2016incorporating} for end-to-end dialogue task completion~\cite{lei2018sequicity,zhang2019task}. 
 Such modules are usually absent in the pre-training stage. Therefore, tasks-specific architecture modifications are required in order to adapt pre-trained language models to different dialogue tasks. 
 
In this work, we aim to simplify the process of transferring the prior knowledge of pre-trained language models for improving task-oriented dialogue systems. 
We propose Minimalist Transfer Learning (\textit{MinTL}), a simple yet effective transfer learning framework that allows to plug-and-play pre-trained sequence-to-sequence (Seq2Seq) models and jointly learn dialogue state tracking (DST) and dialogue response generation. Unlike previous approaches~\cite{lei2018sequicity,zhang2019task}, which use a copy mechanism to ``carryover'' the previous dialogue states and generate new dialogue states, we introduce \textit{Levenshtein belief spans} ($Lev$) which models the difference between old states and new states. In practice, \textit{MinTL} first decodes the $Lev$ for updating the previous dialogue state; then, the updated state is used to search the external knowledge base; and finally, a response decoder decodes response by conditioning on the dialogue context and knowledge base match result. 

\textit{MinTL} is easy to set up by using different pre-trained seq2seq backbones.  We conduct extensive experiments on both DST and end-to-end dialogue response generation tasks with two pre-trained seq2seq models, such as  T5~\cite{raffel2019exploring} and BART~\cite{lewis2019bart}. The experimental result on a large-scale task-oriented dialogue benchmark MultiWOZ~\cite{budzianowski2018multiwoz,eric2019multiwoz} suggests that our proposed method significantly improves SOTA performance in both the full data and simulated low resource setting. Our contributions are summarized as follows: 

\begin{itemize}
    \item We propose the \textit{MinTL} framework that efficiently leverages pre-trained language models for task-oriented dialogue without any ad hoc module.
    \item We propose the novel $Lev$ for efficiently tracking the dialogue state with the minimal length of generation, which greatly reduces the inference latency.
    \item We instantiate our framework with two different pre-trained backbones, and both of them improve the SOTA results by a large margin.
    \item We demonstrate the robustness of our approach in the low-resource setting. By only using 20\% training data, \textit{MinTL}-based systems achieve competitive results compared to the SOTA. 
\end{itemize}

\begin{figure*}[!t]
    \centering
    \includegraphics[width=0.92\linewidth]{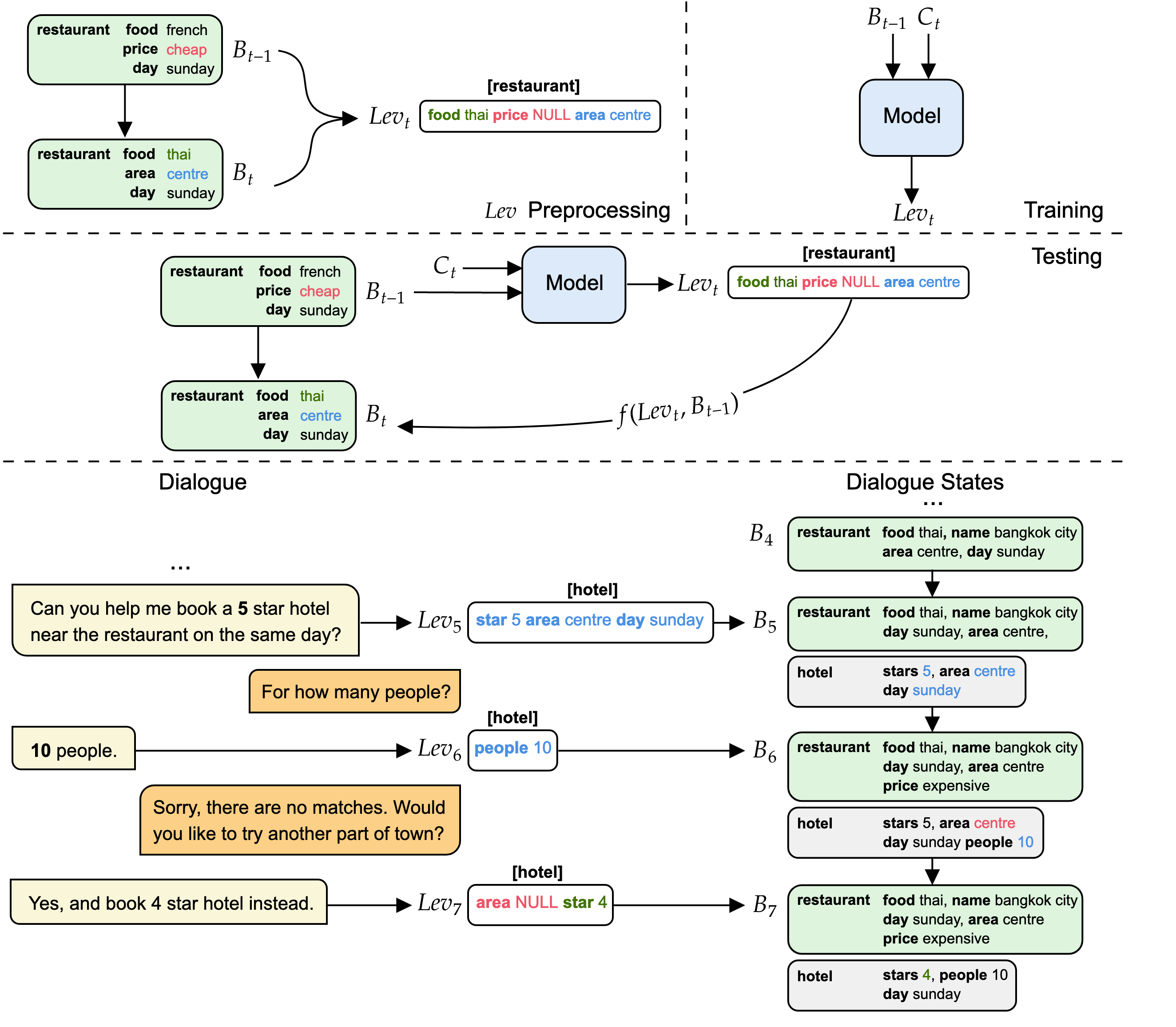}
    \caption{Dialogue state tracking with $Lev$. The model first generates $Lev$, then updates the dialogue state with new generated slot-values. The updating operations are insertion (blue), deletion (red), and substitution (green).}
    \label{fig:Lev}
\end{figure*}

\section{Related Work}
\paragraph{Pre-trained Language Models.} Language model (LM) pre-training ~\cite{radford2019language,devlin2019bert,yang2019xlnet}, 
has been shown to be beneficial in NLP downstream tasks. Generative pre-trained unidirectional LMs (e.g., GPT2) are effective in language generation tasks~\cite{radford2019language,hosseini2020simple,peng2020soloist,lin2020exploring}. Several works have applied a generative pre-training approach in open domain chitchat tasks~\cite{wolf2019transfertransfo,zhang2019dialogpt}, and achieved promising results. On the other hand, bidirectional pre-trained LMs~\cite{devlin2019bert, liu2019roberta} significantly improve the performance of natural language understanding tasks. These models are usually evaluated on classification tasks such as the GLUE benchmark~\cite{wang2018glue}, extractive question answering tasks~\cite{rajpurkar2016squad}, and dialogue context understanding~\cite{wu2020tod}. However, their bidirectionality nature makes them difficult to be applied to natural language generation tasks~\cite{dong2019unified}.
Recent works~\cite{dong2019unified,raffel2019exploring,lewis2019bart} unified unidirectional LM and bidirectional LM pre-training approaches, and proposed a Seq2Seq LM, which are pre-trained with language denoising objectives. A systematic study conducted by~\citet{raffel2019exploring} suggests that the combination of an encoder-decoder architecture and language denoising pre-training objectives yields the best result in both language understanding and generation tasks. Notably, the two latest pre-trained chatbots, Meena~\cite{adiwardana2020towards} and BST~\cite{roller2020recipes}, are also built on an encoder-decoder architecture.
In this work, we transfer the prior knowledge of Seq2Seq LMs to task-oriented dialogues, and successfully improve the SOTA~\cite{zhang2019task} result with less human annotation.

\paragraph{Task-Oriented Dialogue.} 
Task-oriented dialogue systems are designed to accomplish a goal described by a user in natural language. Such systems are usually built with a pipeline approach. The pipeline often requires natural language understanding (NLU) for belief state tracking, dialogue management (DM) for deciding which actions to take, and natural language generation (NLG) for generating responses~\cite{williams2007partially}. To simplify the system design and reduce human supervision, several end-to-end trainable systems have been proposed~\cite{bordes2016learning,wen2017network,lei2018sequicity,neelakantan2019neural,eric-manning:2017:EACLshort,ericKVR2017,madotto2018mem2seq}. These methods have been shown to achieve promising results in single-domain tasks. However, the recently proposed multi-domain task-oriented dialogue datasets~\cite{budzianowski2018multiwoz,eric2019multiwoz} 
bring new challenges for multi-domain dialogue state tracking and response generation. Several follow up works~\cite{wu2019transferable,  chen2019semantically,budzianowski2019hello,mehri2019structured,madotto2020attention} improved on the initial baselines with various methodologies. \citet{zhang2019task} proposed the domain aware multi-decoder network and augmented the system act labels by leveraging the user act annotation, achieving the SOTA results in MultiWoz. However, the aforementioned works rely on task-specific design and extensive human annotations. To reduce the human effort and simplify the system design, we propose a simple transfer learning framework that can be easily set up with pre-trained Seq2Seq models and obtain decent performance with a small fraction of the training data.

\section{Methodology}
In this section, we first provide
the notations that are used throughout the paper, then we introduce the $Lev$ for efficient DST, and finally, describe the \textit{MinTL} framework and two backbone models.

\paragraph{Notations.} Let us define a dialogue $\mathcal{C}=\{U_1,R_1, \dots, U_T, R_T\}$ as an alternating set of utterances from two speakers, where $U$ and $R$ represent the user utterance and the system response, respectively. At turn $t$, we denote a dialogue context as $C_t=\{U_{t-w},R_{t-w}, \dots, R_{t-1},U_t\}$ and system response as $R_t$, where $w$ is the context window size.
$\mathcal{B} = \{B_1,\dots,B_T\}$ is the dialogue states for each turn.  We define $B_t$, the dialogue state at turn $t$, as a dictionary that maps (domain: $d_i$, slot: $s_j$) a pair into values $v$, where $\mathcal{D}=\{d_1,\dots,d_N\}$ are the domains, and $\mathcal{S}=\{s_1,\dots,s_M\}$ are slots to track. Thoughtout the paper, we denote the value of a pair $(d_i,s_j)$ in $B_t$ as $B_t(d_i,s_j) = v$, and $B_t(d_i,s_j) = \varepsilon$ when key $(d_i,s_j)$ is not in $B_t$, where $\varepsilon$ denotes an empty string, and $|\varepsilon|=0$.

\begin{figure*}[!t]
    \begin{minipage}{.5\linewidth}
    \centering
    \tikzset{every picture/.style={line width=0.75pt}} 

\begin{tikzpicture}[x=0.75pt,y=0.75pt,yscale=-1,xscale=1]

\draw  [fill={rgb, 255:red, 155; green, 155; blue, 155 }  ,fill opacity=0.11 ] (91,264.6) .. controls (91,259.85) and (94.85,256) .. (99.6,256) -- (128.33,256) .. controls (133.08,256) and (136.93,259.85) .. (136.93,264.6) -- (136.93,290.4) .. controls (136.93,295.15) and (133.08,299) .. (128.33,299) -- (99.6,299) .. controls (94.85,299) and (91,295.15) .. (91,290.4) -- cycle ;
\draw    (61.5,288) -- (86,288) ;
\draw [shift={(89,288)}, rotate = 180] [fill={rgb, 255:red, 0; green, 0; blue, 0 }  ][line width=0.08]  [draw opacity=0] (8.93,-4.29) -- (0,0) -- (8.93,4.29) -- cycle    ;
\draw    (255.67,301.67) -- (255.56,324.37) ;
\draw [shift={(255.54,327.37)}, rotate = 270.27] [fill={rgb, 255:red, 0; green, 0; blue, 0 }  ][line width=0.08]  [draw opacity=0] (8.93,-4.29) -- (0,0) -- (8.93,4.29) -- cycle    ;
\draw    (62.5,268) -- (87,268) ;
\draw [shift={(90,268)}, rotate = 180] [fill={rgb, 255:red, 0; green, 0; blue, 0 }  ][line width=0.08]  [draw opacity=0] (8.93,-4.29) -- (0,0) -- (8.93,4.29) -- cycle    ;
\draw  [fill={rgb, 255:red, 155; green, 155; blue, 155 }  ,fill opacity=0.11 ] (233,264.6) .. controls (233,259.85) and (236.85,256) .. (241.6,256) -- (270.33,256) .. controls (275.08,256) and (278.93,259.85) .. (278.93,264.6) -- (278.93,290.4) .. controls (278.93,295.15) and (275.08,299) .. (270.33,299) -- (241.6,299) .. controls (236.85,299) and (233,295.15) .. (233,290.4) -- cycle ;
\draw    (143,278.07) -- (226,278) ;
\draw [shift={(229,278)}, rotate = 539.95] [fill={rgb, 255:red, 0; green, 0; blue, 0 }  ][line width=0.08]  [draw opacity=0] (8.93,-4.29) -- (0,0) -- (8.93,4.29) -- cycle    ;
\draw  [fill={rgb, 255:red, 155; green, 155; blue, 155 }  ,fill opacity=0.11 ] (91,387.6) .. controls (91,382.85) and (94.85,379) .. (99.6,379) -- (128.33,379) .. controls (133.08,379) and (136.93,382.85) .. (136.93,387.6) -- (136.93,413.4) .. controls (136.93,418.15) and (133.08,422) .. (128.33,422) -- (99.6,422) .. controls (94.85,422) and (91,418.15) .. (91,413.4) -- cycle ;
\draw    (113.93,303.57) -- (113.93,372) ;
\draw [shift={(113.93,375)}, rotate = 270] [fill={rgb, 255:red, 0; green, 0; blue, 0 }  ][line width=0.08]  [draw opacity=0] (8.93,-4.29) -- (0,0) -- (8.93,4.29) -- cycle    ;
\draw    (86.41,399.17) -- (64.41,399.17) ;
\draw [shift={(61.41,399.17)}, rotate = 360] [fill={rgb, 255:red, 0; green, 0; blue, 0 }  ][line width=0.08]  [draw opacity=0] (8.93,-4.29) -- (0,0) -- (8.93,4.29) -- cycle    ;
\draw   (272,387.78) -- (272,409.23) .. controls (272,412.41) and (264.6,415) .. (255.46,415) .. controls (246.33,415) and (238.93,412.41) .. (238.93,409.23) -- (238.93,387.78)(272,387.78) .. controls (272,390.96) and (264.6,393.55) .. (255.46,393.55) .. controls (246.33,393.55) and (238.93,390.96) .. (238.93,387.78) .. controls (238.93,384.59) and (246.33,382) .. (255.46,382) .. controls (264.6,382) and (272,384.59) .. (272,387.78) -- cycle ;
\draw    (255.59,354.29) -- (255.48,377) ;
\draw [shift={(255.46,380)}, rotate = 270.27] [fill={rgb, 255:red, 0; green, 0; blue, 0 }  ][line width=0.08]  [draw opacity=0] (8.93,-4.29) -- (0,0) -- (8.93,4.29) -- cycle    ;
\draw    (231.93,399) -- (145,399.48) ;
\draw [shift={(142,399.5)}, rotate = 359.68] [fill={rgb, 255:red, 0; green, 0; blue, 0 }  ][line width=0.08]  [draw opacity=0] (8.93,-4.29) -- (0,0) -- (8.93,4.29) -- cycle    ;

\draw (21,254.5) node [anchor=north west][inner sep=0.75pt]  [font=\normalsize]  {$\mathcolorbox{soulred}{B_{t-1}}$};
\draw (97,270) node [anchor=north west][inner sep=0.75pt]  [font=\large]  {$\mathbf{Enc}$};
\draw (185,329) node [anchor=north west][inner sep=0.75pt]  [font=\normalsize]  {$B_{t} =f( \mathcolorbox{soulbleu}{Lev_{t}} ,B_{t-1})$};
\draw (36,276.5) node [anchor=north west][inner sep=0.75pt]  [font=\normalsize]  {$\mathcolorbox{soulgreen}{C_{t}}$};
\draw  [color={rgb, 255:red, 255; green, 255; blue, 255 }  ,draw opacity=1 ][fill={rgb, 255:red, 255; green, 255; blue, 255 }  ,fill opacity=1 ]  (170,265) -- (191,265) -- (191,289) -- (170,289) -- cycle  ;
\draw (173,270) node [anchor=north west][inner sep=0.75pt]  [font=\normalsize]  {$H$};
\draw (235,268.6) node [anchor=north west][inner sep=0.75pt]  [font=\large]  {$\mathbf{Dec}_{L}$};
\draw  [color={rgb, 255:red, 255; green, 255; blue, 255 }  ,draw opacity=1 ][fill={rgb, 255:red, 255; green, 255; blue, 255 }  ,fill opacity=1 ]  (103,323) -- (124,323) -- (124,347) -- (103,347) -- cycle  ;
\draw (106,329) node [anchor=north west][inner sep=0.75pt]  [font=\normalsize]  {$H$};
\draw (93,391.6) node [anchor=north west][inner sep=0.75pt]  [font=\large]  {$\mathbf{Dec}_{R}$};
\draw (35,386) node [anchor=north west][inner sep=0.75pt]  [font=\normalsize]  {$\mathcolorbox{soulorange}{R_{t}}$};
\draw (244.93,396.78) node [anchor=north west][inner sep=0.75pt]  [font=\normalsize]  {\texttt{KB}};
\draw  [draw opacity=0][fill={rgb, 255:red, 255; green, 255; blue, 255 }  ,fill opacity=1 ]  (181.9,386.57) -- (202.9,386.57) -- (202.9,411.57) -- (181.9,411.57) -- cycle  ;
\draw (180.9,387.57) node [anchor=north west][inner sep=0.75pt]  [font=\normalsize]  {$\mathcolorbox{soulyellow}{k_{t}}$};
\end{tikzpicture}
    \end{minipage}%
    \begin{minipage}{.5\linewidth}
    \begin{tabular}{l}
    \begin{tabular}[c]{@{}l@{}} \hlred{[hotel] stars 5 area centre day sunday [restaurant]} \\ \hlred{food thai area centre day sunday name bangkok} \\ \hlred{ city \textless EOB\textgreater} \hlgreen{Can you help me book a 5 star} \\ \hlgreen{hotel near the restaurant on the same day?  } \\ \hlgreen{\textless EOU\textgreater For how many people? \textless EOR\textgreater 10} \\ \hlgreen{ people \textless EOU\textgreater} \end{tabular} \\ \hline\hline
    \hlblue{\textless SOB\textgreater [hotel] people 10 \textless EOB\textgreater} \\ \hline\hline
    \begin{tabular}[c]{@{}l@{}}\hlyellow{\textless KB2\textgreater} \hlorange{sorry, there are no matches. would you} \\ \hlorange{like to try another part of town? \textless EOR\textgreater}\end{tabular}
    \end{tabular}
    \end{minipage}%
    \caption{Overview of the \textit{MinTL} framework. The left figure shows the information flow among all modules. The explicit inputs and outputs of each module are described on the right. \textit{MinTL} first encodes previous dialogue state $B_t$ and dialogue context $C_t$, and decodes $Lev_t$. Then $Lev_t$ is used to update $B_{t-1}$ to $B_{t}$ via function $f$. The updated $B_t$ is used to query the KB and booking API and return KB state $k_t$. Finally, the $R_t$ is generated by conditioning on $B_{t-1}$, $C_t$ and $k_t$.}
    \label{fig:minimalist}
\end{figure*}

\subsection{Levenshtein Belief Spans}
The goal of DST is to track the slot values for each domain mentioned in dialogue. Existing works either perform classifications for each slot over a candidate-value list~\cite{zhang2019find} or directly generate slot values with a generative model~\cite{lei2018sequicity, wu2019transferable,kim2019efficient,Le2020Non-Autoregressive}. Notably, \citet{lei2018sequicity} introduce the concept of \textit{Belief span} that reformats the dialogue states into a text span for allowing models to generate slot values dynamically. Compared to classification based DST, generative DST models can predict the slot values without full access to predefined ontology. However, the aforementioned generative methods either generate the belief span from scratch~\cite{lei2018sequicity} or classify the state operations over all the combinations of domain slot pairs for decoding necessary slot values~\cite{kim2019efficient,Le2020Non-Autoregressive}, which is not scalable when interfacing to a large number of services and APIs spanning multiple domains~\cite{rastogi2019towards}. 

The idea of $Lev$ is to generate minimal belief spans at each turn for editing the previous dialogue states. As illustrated in Figure~\ref{fig:Lev}, $Lev$ is constructed at training time as the DST training target. 
Given $B_{t-1}$, $B_t$, and a pair of $(d_i,s_j)$, we define the three slot level edit operation conditions, i.e., insertion (\texttt{INS}), deletion (\texttt{DEL}) and substitution (\texttt{SUB}), as:
\begin{align}
    \texttt{INS} &\rightarrow B_t(d_i,s_j) \neq \varepsilon \land B_{t-1}(d_i,s_j) = \varepsilon \\
    \texttt{DEL} &\rightarrow B_t(d_i,s_j) = \varepsilon \land B_{t-1}(d_i,s_j) \neq \varepsilon \\
    \texttt{SUB} &\rightarrow B_t(d_i,s_j) \neq B_{t-1}(d_i,s_j).
\end{align}
In domain $d_i$, to update the $B_{t-1}(d_i,s_j)$ to $B_t(d_i,s_j)$, the minimal slot-value pair needed to be generated is $E(d_i,s_j)$, defined as
\begin{equation}
    E(d_i,s_j) = 
    \begin{cases}
      s_j \oplus B_t(d_i,s_j) & \text{if}\ \texttt{INS} \\ 
      s_j \oplus \texttt{NULL} & \text{if}\ \texttt{DEL} \\ 
      s_j \oplus B_t(d_i,s_j) & \text{if}\ \texttt{SUB} \\ 
      \varepsilon & \text{otherwise}, \\
    \end{cases} 
\end{equation}
where $\oplus$ denotes string concatenation. \texttt{NULL} is the symbol denoting to delete the slot $(d_i,s_j)$ from $B_{t-1}$. Then, we aggregate all the $E(d_i,s_j)$ for domain $d_i$ as follows:
\begin{equation}
L(d_i) = E(d_i,s_1) \oplus \cdots \oplus E(d_i,s_M).
\end{equation}
When the dialogue state of domain $d_i$ needs to be updated, i.e., $L(d_i) \neq \varepsilon$, we append the domain information $[d_i]$ at the beginning of $L(d_i)$ to construct $Lev$ of domain $d_i$:
\begin{equation}
\delta(L,d_i) = 
    \begin{cases}
      [d_i] \oplus  L(d_i) & \text{if}\  L(d_i) \neq \varepsilon \\ 
      \varepsilon & \text{otherwise}. \\
    \end{cases}
\end{equation}
Finally, we formally define $Lev$ as the following:
\begin{equation}
Lev = \delta(L,d_1) \oplus \cdots \oplus \delta(L,d_N).
\end{equation}

At inference time, the model first generates $Lev_t$ at turn $t$, then edits the $B_{t-1}$ by using a deterministic function $f$, defined as:
\begin{equation}
B_t = f(Lev_t, B_{t-1}).
\label{edit_func}
\end{equation}
This function simply update the $B_{t-1}$ when new slot-value pairs appear in $Lev_t$, and it delete the corresponding slot-value when the \texttt{NULL} symbol is generated.

Figure~\ref{fig:Lev} shows an example of editing the dialogue state editing process using $Lev$. In the $6$-th turn, the generated $Lev_6$ inserts the value $10$ into the slot \textit{people}. In the $7$-th turn, the $\texttt{NULL}$ in $Lev_7$ triggers the \texttt{DEL} operation, and thus the slot $(hotel,area)$ is deleted in $B_{6}$, which is equivalent to $B_7(hotel,area)=\varepsilon$.

\subsection{\textit{MinTL} Framework}

Figure \ref{fig:minimalist} describes the flow of the \textit{MinTL} framework with a general encoder-decoder architecture.
 The input of our framework is a dialogue context $C_t$ and a previous dialogue state $B_{t-1}$. All sub-sequences are concatenated with special segment tokens, i.e., $B_{t-1}$\textless $EOB$\textgreater \dots $R_{t-1}$\textless $EOR$\textgreater $U_t$\textless $EOU$\textgreater, as input to the encoder.

\begin{equation}
H = Encoder(\mathcal{C}_t, B_{t-1}),
\end{equation}
where the $H \in \mathbb{R}^{I \times d_{model}}$ is the hidden states of the encoder, and $I$ is the input sequence length. Then, the $Lev$ decoder attends to the encoder hidden states $H$ and decodes $Lev_t$ sequentially:
\begin{equation}
Lev_t= Decoder_L(H).
\end{equation}
The learning objective of this generation process is minimizing the negative log-likelihood of $Lev_t$ given $\mathcal{C}_t$ and $B_{t-1}$, that is
\begin{equation}
\mathcal{L}_{L} = -\log p(Lev_t | \mathcal{C}_t, B_{t-1}).
\end{equation}
The generated $Lev_t$ is used for editing the $B_{t-1}$ with the deterministic function $f$ described in Equation \ref{edit_func}.

The updated $B_t$ is used to query the external knowledge (KB) and booking APIs. We first categorize the query result $k_t$ according to the number of matching entities and the booking availability (a detailed list of $k_t$ values is provided in the Appendix A). According to the result, we look up one embedding $e_{k}\in \mathbb{R}^{d_{model}}$ from the set of learnable KB state embeddings $E_{k}\in \mathbb{R}^{K \times d_{model}}$~\footnote{KB state embeddings can be easily constructed by extending token embeddings of pre-trained models.}, where $K$ is the number of possible KB states. Then, the looked up embedding $e_{k}$ is used as the starting token embedding of the response decoder for generating the delexicalized response $R_t$:

\begin{equation}
R_t= Decoder_R(H, e_{k}).
\end{equation}
The learning objective of response generation is minimizing the negative log-likelihood of $R_t$ given $B_{t-1}$, $\mathcal{C}_t$ and $k_t$,
\begin{equation}
\mathcal{L}_{R}= -\log p(R_t | \mathcal{C}_t, B_{t-1}, k_t).
\end{equation}
Different from previous works~\cite{lei2018sequicity, zhang2019task}, our response generation process is not condition on $B_t$ because the dialogue context $C_t$ already includes the information of $B_t$.

During training, all parameters are jointly optimized by minimizing the sum of the $Lev$ generation and response generation losses:
\begin{equation}
\mathcal{L} = \mathcal{L}_{L} + \mathcal{L}_{R}.
\end{equation}

\subsection{Backbone Models}
Our framework can be easily set up with pre-trained language models by initializing the encoder and decoders with pre-trained weights. We briefly introduce the two pre-trained backbones used in this paper: BART~\cite{lewis2019bart} and Text-To-Text Transfer Transformer (T5)~\cite{raffel2019exploring}.

\paragraph{BART} is implemented as a standard encoder-decoder Transformer with a bidirectional encoder and an autoregressive decoder. It is pre-trained as denoising autoencoders which corrupt documents, and then optimize a reconstruction loss—the cross-entropy between the decoder’s output and the original document. BART applies five different document corruption methods in the pre-training, including Token Masking~\cite{devlin2019bert}, Token Deletion, Text Infilling~\cite{joshi2020spanbert}, Sentence Permutation, and Document Rotation. 
\paragraph{T5} is an encoder-decoder Transformer with relative position embeddings~\cite{shaw2018self}. The model is pre-trained on the Colossal Clean Crawled Corpus (C4)~\cite{raffel2019exploring} that contains about 750GB of clean and natural English text. The pre-training objective is spans prediction, i.e., masking out 15\% of input spans, then predicting the missing spans using the decoder.

\section{Experiments}

\begin{table*}[!t]
\resizebox{0.99\textwidth}{!}{
\begin{tabular}{@{}lccccccc@{}}
\toprule
\multirow{2}{*}{\textbf{Model}} & \multicolumn{3}{c}{\textbf{Supervision}} & \multirow{2}{*}{\textbf{Inform (\%)}} & \multirow{2}{*}{\textbf{Success (\%)}} & \multirow{2}{*}{\textbf{BLEU}} & \multirow{2}{*}{\textbf{Combined}} \\ \cmidrule(lr){2-4}
 & Dialogue State & System Act & User Act & & & & \\ \midrule
\multicolumn{1}{l|}{Seq2Seq$^\star$} & \textbf{oracle} & \xmark & \multicolumn{1}{c|}{\xmark} & 76.70 & 64.63 & 18.05 & 88.72 \\
\multicolumn{1}{l|}{GPT2-small$^\star$} & \textbf{oracle} & \xmark & \multicolumn{1}{c|}{\xmark} & 66.43 & 55.16 & 18.02 & 78.82 \\
\multicolumn{1}{l|}{GPT2-medium$^\star$} & \textbf{oracle} & \xmark & \multicolumn{1}{c|}{\xmark} & 70.96 & 61.36 & 19.05 & 85.21 \\ 
\multicolumn{1}{l|}{MD-Sequicity} & \cmark & \xmark & \multicolumn{1}{c|}{\xmark} & 75.72 & 58.32 & 15.40 & 82.40 \\
\multicolumn{1}{l|}{HRED-TS$^\star$} & \cmark & \cmark& \multicolumn{1}{c|}{\xmark} & 70.00 & 58.00 & 17.50 & 81.50 \\
\multicolumn{1}{l|}{SFN + RL$^\star$} & \cmark & \cmark& \multicolumn{1}{c|}{\xmark} & 73.80 & 58.60 & 18.27 & 84.47 \\
\multicolumn{1}{l|}{DAMD} & \cmark & \cmark& \multicolumn{1}{c|}{\xmark} & 72.79 & 60.43 & 16.93 & 83.54 \\
\multicolumn{1}{l|}{DAMD + multi-action} & \cmark & \cmark& \multicolumn{1}{c|}{\cmark} & 76.33 & 64.25 & 17.96 & 88.25 \\
\multicolumn{1}{l|}{Sequicity (T5-small)} & \cmark & \xmark & \multicolumn{1}{c|}{\xmark} & 71.64 & 61.01 & 18.02 & 84.35 \\
\multicolumn{1}{l|}{\textit{MinTL} (T5-small)} & \cmark & \xmark & \multicolumn{1}{c|}{\xmark} & 80.04 & 72.71 & \textbf{19.11} & 95.49 \\
\multicolumn{1}{l|}{\textit{MinTL} (T5-base)} & \cmark & \xmark & \multicolumn{1}{c|}{\xmark} & 82.15 & 74.44 & 18.59 & 96.88 \\
\multicolumn{1}{l|}{\textit{MinTL} (BART-large)} & \cmark & \xmark & \multicolumn{1}{c|}{\xmark} & \textbf{84.88} & \textbf{74.91} & 17.89 & \textbf{97.78} \\ \bottomrule
\end{tabular}
}
\caption{End-to-end response generation results on MultiWOZ2.0. \cmark and \xmark $ $ denote whether a model leverages dialogue state, and/or speech act annotations during training. \textbf{oracle} denotes the gold dialogue state is used in both training and test time. Our results are averaged over three random seeds. $^\star$: results reported by the original paper.}
\label{end2end}
\end{table*}

\subsection{Datasets}
We evaluate the proposed framework on the MultiWOZ dataset. It is a large-scale multi-domain task-oriented dialogue benchmark collected via the Wizard-of-Oz setting. The dataset contains 8438/1000/1000 dialogues for training/validation/testing, respectively. The dialogues in the corpus span over seven domains (restaurant, train, attraction, hotel, taxi, hospital, and police), and each dialogue session contains one to three domains. There are two existing dataset versions: MultiWOZ 2.0~\cite{budzianowski2018multiwoz} and MultiWOZ 2.1~\cite{eric2019multiwoz}. We test the dialogue state tracking module of our framework on both datasets, and end-to-end models on MultiWOZ 2.0.

\subsection{Implementation Details}
We set up our framework with three pre-trained models: 1) T5-small (60M parameters) has 6 encoder-decoder layers and each layer has 8-headed attention with hidden size $d_{model}=512$; 2) T5-base (220M parameters) has 12 encoder-decoder layers, and each of them has 12-headed attention with hidden size $d_{model}=768$;  3) BART-large (400M parameters) has 12 encoder-decoder layers, each layer has 16-headed attention with hidden size $d_{model}=1024$. We add special segment token embeddings and KB state embeddings to pre-trained models by extending the token embeddings. For a fair comparison, we use the pre-processing script released by~\citet{zhang2019task}~\footnote{https://gitlab.com/ucdavisnlp/damd-multiwoz}. All the models are fine-tuned with a batch size of 64 and early stop according to the performance on the validation set. Our implementation is based on HuggingFace Transformers library~\cite{Wolf2019HuggingFacesTS}. We report the training hyper-parameters of each model in Appendix B.

\subsection{Evaluation Metrics}
For the end-to-end dialogue modeling task, there are three automatic metrics to evaluate the response quality: 1) \textbf{Inform} rate: if the system provides a correct entity, 2) \textbf{Success} rate: if the system provides the correct entity and answers all the requested information, 3) \textbf{BLEU} \cite{papineni2002bleu} for measuring the fluency of the generated response. Following previous work~\cite{mehri2019structured}, we also report the combined score, i.e., \textbf{Combined} = (Inform + Success)$\times 0.5$ + BLEU, as an overall quality measure. Joint goal accuracy (\textbf{Joint Acc.}) is used to evaluate the performance of the DST. The model outputs are only counted as correct when all of the predicted values exactly match the oracle values.

\begin{table*}[!t]
\resizebox{0.98\textwidth}{!}{
\begin{tabular}{@{}lccccccccc@{}}
\toprule
\multirow{2}{*}{\textbf{ Model}}            & \multicolumn{3}{c}{\textit{\textbf{5\%}}}     & \multicolumn{3}{c}{\textit{\textbf{10\%}}}    & \multicolumn{3}{c}{\textit{\textbf{20\%}}} \\ 
                                           & \textbf{Inform} & \textbf{Success} & \textbf{BLEU}  & \textbf{Inform} & \textbf{Success} & \textbf{BLEU}  & \textbf{Inform}       & \textbf{Success}       & \textbf{BLEU}        \\ \midrule
\multicolumn{1}{l|}{MD-Sequicity}          & 49.40  & 19.70   & \multicolumn{1}{c|}{10.30} & 58.10  & 34.70   & \multicolumn{1}{c|}{11.40} & 64.40        & 42.10         & 13.00       \\
\multicolumn{1}{l|}{DAMD}                  & 57.20  & 27.00   & \multicolumn{1}{c|}{9.90}  & 58.30  & 33.90   & \multicolumn{1}{c|}{13.30} & 67.40        & 40.10         & 13.80       \\
\multicolumn{1}{l|}{DAMD + multi-action}   & 56.60  & 24.50   & \multicolumn{1}{c|}{10.60} & 62.00  & 39.40   & \multicolumn{1}{c|}{14.50} & 68.30        & 42.90         & 11.80       \\
\multicolumn{1}{l|}{\textit{MinTL} (T5-small)} & 58.86  & 49.35   & \multicolumn{1}{c|}{\textbf{14.51}} & 63.16  & 52.65   & \multicolumn{1}{c|}{\textbf{15.71}} & 73.57        & 66.07         & \textbf{17.55}       \\
\multicolumn{1}{l|}{\textit{MinTL} (T5-base)}  & 69.57  & 57.76   & \multicolumn{1}{c|}{14.50} & 72.17  & 61.16   & \multicolumn{1}{c|}{15.56} & 78.98        & \textbf{70.37}         & 16.69       \\
\multicolumn{1}{l|}{\textit{MinTL} (BART-large)}                    & \textbf{75.48}  & \textbf{60.96}   & \multicolumn{1}{c|}{13.98}                      & \textbf{78.08}  & \textbf{66.87}   & \multicolumn{1}{c|}{15.46}                      & \textbf{82.48}        & 68.57         & 13.00       \\ \bottomrule
\end{tabular}
}
\caption{Results of simulated low resource experiments. 5\% (400 dialogues), 10\% (800 dialogues), 20\% (1600 dialogues) of training data is used to train each model.}
\label{few_shot}
\end{table*}

\begin{table}[!t]
\resizebox{0.49\textwidth}{!}{
\begin{tabular}{@{}lccc@{}}
\toprule
\textbf{Model}                & \textbf{Inform (\%)} & \textbf{Success (\%)} & \textbf{BLEU}  \\ \midrule
\multicolumn{1}{l}{\textit{MinTL} (T5-small)} & \textbf{80.04}                         & \textbf{72.71}                          & 19.11                   \\
\multicolumn{1}{l}{w/o Lev}  & 71.62                         & 63.20                          & 16.11                                    \\
\multicolumn{1}{l}{w/ shared decoder}             & 74.90                         & 67.03                          & \textbf{20.10}                                    \\ \bottomrule
\end{tabular}
}
\caption{Ablation study on different variants of \textit{MinTL} on MultiWOZ 2.0 in the end-to-end evaluation setting.}
\label{ablation}
\end{table}

\subsection{Baselines}
\subsubsection{End-to-end Modeling}

\paragraph{Oracle DST:} \textbf{Seq2Seq}, fine-tuned \textbf{GPT2-small}, and \textbf{GPT2-medium}~\cite{radford2019language} with oracle dialogue state as input~\cite{budzianowski2018multiwoz}.
\paragraph{HRED-TS:} a teacher-student framework with a hierarchical recurrent encoder-decoder backbone~\cite{peng2019teacher}.
\paragraph{SFN + RL:} a seq2seq network comprised of several pre-trained dialogue modules that are connected through hidden states. Reinforcement fine tuning is used additionally to train the model~\cite{mehri2019structured}.
\paragraph{MD-Sequicity:} an extension of the \textit{Sequicity}~\cite{lei2018sequicity} framework for multi-domain task-oriented dialogue by~\citet{zhang2019task}.
\paragraph{DAMD:} the domain-aware multi-decoder network proposed by ~\citet{zhang2019task}. The author also proposed the multi-action data augmentation method by leveraging system act and user act annotations. We denote the method as \textbf{DAMD + multi-action}.
\paragraph{Sequicity + T5:} The \textit{Sequicity}~\cite{lei2018sequicity} framework with the T5 backbone model~\cite{raffel2019exploring}. There are two main differences between \textit{Sequicity} and our framework: 1) \textit{Sequicity} generates dialogue states from scratch at each turn, 2) \textit{MinTL} generates responses by conditioning on dialogue context $C_t$ instead of new generated dialogue state $B_t$.

\subsubsection{Dialogue State Tracking}
We compare our DST module with both the classification-based DST and generation-based DST baselines. The former includes MDBT~\cite{ramadan2018large}, GLAD~\cite{zhong2018global}, GCE~\cite{nouri2018toward}, FJST \cite{eric2019multiwoz}, HyST \cite{goel2019hyst}, SUMBT~\cite{lee2019sumbt}, SST \cite{Chen2020SchemaGuidedMD}, TOD-BERT~\cite{wu2020tod}, and DST-Picklist \cite{zhang2019find}; the latter includes Neural Reading \cite{gao2019dialog},
TRADE \cite{wu2019transferable},
COMER \cite{ren2019scalable},
SOM-DST \cite{kim2019efficient},
DSTQA \cite{zhou2019multi}, and
NADST \cite{Le2020Non-Autoregressive}. 

\subsection{Results}
\subsubsection{End-to-end Modeling}
We first compare our systems with baselines in the end-to-end dialogue learning setting, where the generated dialogue states are used for the knowledge base search and response generation. The results are shown in Table~\ref{end2end}. \textit{MinTL}-based systems achieve the best performance in terms of inform rate, success rate, and BLEU. With fewer human annotations, our models improve the previous SOTA model~\cite{zhang2019task}  by around a 10\% success rate. Using T5-small as the backbone barely improves the overall performance of \textit{Sequicity}~\cite{lei2018sequicity}, because the copy mechanism~\cite{gu2016incorporating} is absent in this pre-trained model. Compared to the \textit{Sequicity} framework, our approach achieves an around 11\% higher success rate with the same backbone model, which suggests that \textit{MinTL} is able to effectively leverage pre-trained language models.

\paragraph{Low Resource Settings.}
We evaluate our models in the simulated low resource setting to test if transferring a pre-trained language model to task-oriented dialogue can alleviate the data scarcity problem. Specifically, we use 5\%, 10\%, and 20\% of the training set data to train our models and baselines. The result is reported in Table~\ref{few_shot}. \textit{MinTL}-based systems consistently outperform the DAMD~\cite{zhang2019task}, MD-Sequicity~\cite{lei2018sequicity} baselines by a large margin, which demonstrates the effectiveness of transfer learning. It is worth noting that the performance gap between \textit{MinTL} and baselines decreases with respect to the increase in the training data size. This indicates that prior knowledge from the pre-trained language model is more important in the extremely low-resource scenarios. With only 20\% of training data, our models can achieve competitive results compared to the full data trained DAMD model.

\paragraph{Ablation Study.}
We conduct a simple ablation study with the T5-small backbone to understand the different variants of \textit{MinTL}. We test our framework with: 1) the belief span proposed by \citet{lei2018sequicity}, and 2) sharing the decoder parameter for both $Lev$ generation and response generation. The result is reported in Table~\ref{ablation}. Replacing $Lev$ with \textit{belief span} hurts the overall performance, which shows the effectiveness of $Lev$. In section \ref{sec:DST}, we also show that $Lev$ greatly reduces the inference latency. On the other hand, although the $Lev$ generation and response generation are conditioned on different starting tokens, sharing the parameters of the two decoders decreases both inform and success rate. It is important to decouple the two decoders because the distributions between the $Lev$ decoder and response decoder are different.

\begin{table}[!t]
\resizebox{0.49\textwidth}{!}{
\begin{tabular}{@{}lcc@{}}
\toprule

\multirow{2}{*}{\textbf{Model}} & \multicolumn{2}{c}{\textbf{MWoZ Joint Acc.}} \\ \cmidrule{2-3}                                                                        & \multicolumn{1}{c}{\textbf{2.0}} & \multicolumn{1}{c}{\textbf{2.1}} \\ \midrule
\multicolumn{1}{l}{MDBT \cite{ramadan2018large}{$^\dagger$}}        & 15.57                                              & -                                                  \\
\multicolumn{1}{l}{GLAD \cite{zhong2018global}{$^\dagger$}}         & 35.57                                              & -                                                  \\
\multicolumn{1}{l}{GCE \cite{nouri2018toward}{$^\dagger$}}          & 36.27                                              & -                                                  \\
\multicolumn{1}{l}{FJST \cite{eric2019multiwoz}{$^\star$}}          & 40.20                                              & 38.00                                               \\
\multicolumn{1}{l}{HyST \cite{goel2019hyst}{$^\dagger$}}            & 44.24                                              & -                                                  \\
\multicolumn{1}{l}{SUMBT \cite{lee2019sumbt}{$^\dagger$}}           & 46.65                                              & -                                                  \\
\multicolumn{1}{l}{TOD-BERT \cite{wu2020tod}{$^\star$}}           & -                                              & 48.00                                                  \\
\multicolumn{1}{l}{DST-Picklist \cite{zhang2019find}{$^\star$}}     & -                                                  & 53.30                                              \\
\multicolumn{1}{l}{SST \cite{Chen2020SchemaGuidedMD}{$^\star$}}     & \textbf{51.17}                                              & \textbf{55.23}                                              \\ \midrule
\multicolumn{1}{l}{Neural Reading \cite{gao2019dialog}{$^\dagger$}} & 41.10                                              & -                                                  \\
\multicolumn{1}{l}{TRADE \cite{wu2019transferable}{$^\dagger$}}     & 48.62                                              & 45.60                                               \\
\multicolumn{1}{l}{COMER \cite{ren2019scalable}{$^\dagger$}}        & 48.79                                              & -                                                  \\
\multicolumn{1}{l}{DSTQA \cite{zhou2019multi}{$^\dagger$}}          & 51.44                                              & 51.17                                              \\
\multicolumn{1}{l}{SOM-DST \cite{kim2019efficient}{$^\star$}}       & 51.38                                              & 52.57                                              \\
\multicolumn{1}{l}{NADST \cite{Le2020Non-Autoregressive}{$^\star$}} & 50.52                                              & 49.04                                              \\
\multicolumn{1}{l}{\textit{MinTL} (T5-small)}                                              & 51.24                                              & 50.95                                              \\
\multicolumn{1}{l}{\textit{MinTL} (T5-base)}                                               & 52.07                                              & 52.52                                              \\
\multicolumn{1}{l}{\textit{MinTL} (BART-large)}                                                                 & \textbf{52.10}                    & \textbf{53.62}                                              \\ \bottomrule
\end{tabular}
}
\caption{Dialogue state tracking results on MultiWOZ 2.0 and MultiWOZ 2.1. The upper part and lower part of the table show the joint goal accuracy of the classification-based and generation-based model, respectively. $^{\dagger}$: results reported by \href{https://github.com/budzianowski/multiwoz}{the leaderboard}. 
{$^\star$}: results reported by the original paper.}
\label{dst}
\end{table}

\subsubsection{Dialogue State Tracking}
\label{sec:DST}
Table \ref{dst} reports the DST results on MultiWOZ 2.0 and MultiWOZ 2.1. \textit{MinTL}-based BART model achieves the highest joint goal accuracy among the generation-based DST models on both datasets. Compared to the SOTA classification-based DST model SST~\cite{Chen2020SchemaGuidedMD}, our model obtains a 1.62\% lower joint goal accuracy on MultiWOZ 2.1. This is because classification-based models have the advantage of predicting slot values from valid candidates. However, having one classifier per domain-slot pair is not scalable when the number of slots and values grow~\cite{lei2018sequicity}. In contrast, our model only generates minimal slot-value pairs when necessary. In our error analysis, we found that our model sometimes generates invalid slot values (e.g., \textit{the cambridge punte} instead of \textit{the cambridge punter} for the \textit{taxi-destination} slot), which can be avoided with a full ontology constraint.

\begin{table}[!t]
\resizebox{0.49\textwidth}{!}{
\begin{tabular}{@{}lllll@{}}
\toprule
\textbf{Model}                 & \textbf{Joint Acc} & \textbf{Latency} & \textbf{Speed Up} & \textbf{NoT} \\ \midrule
TRADE$^\star$                 & 45.60      & 362.15  & ×2.12    & -               \\
TSCP$^\star$                  & 37.12     & 767.57  & ×1.00    & -               \\
NADST$^\star$                 & 49.04     & \textbf{27.31}   & ×28.11   & -               \\
Sequicity (T5-small)  & 44.10      & 200.48  & ×3.83    & 20.99           \\
\textit{MinTL} (T5-small) & \textbf{50.95}     & 49.26   & ×15.58   & 6.58           \\ \bottomrule
\end{tabular}
}
\caption{Latency analysis on MultiWOZ 2.1. \textbf{Latency} denotes the average inference time (ms) per turn and \textbf{NoT} denotes the average number of generated tokens per turn. $^\star$: results borrowed from~\citet{Le2020Non-Autoregressive}}
\label{latency}
\end{table}

\paragraph{Latency Analysis.} Table \ref{latency} reports the average inference time (ms) of each model on the test set of MultiWOZ 2.1. Following \citet{Le2020Non-Autoregressive}, we compute the latency of each model on Nvidia V100 with a batch size of 1. Our model is 15 times faster than TSCP~\cite{lei2018sequicity} and around 7 times faster than TRADE~\cite{wu2019transferable}. On the other hand, our model is slower than NADST~\cite{Le2020Non-Autoregressive}, which is explicitly optimized for inference speed using the non-autoregressive decoding strategy. However, it is hard to incorporate NADST into end-to-end response generation models due to its task-specific architecture design (e.g., fertility decoder). Finally, we compare the generative DST modules of two end-to-end models. By using same backbone model, \textit{MinTL} is around 4 times faster than Sequicity by generating only 6 tokens per turn, which suggests that $Lev$ significantly improves the inference efficiency.

\section{Conclusion}
In this paper, we proposed \textit{MinTL}, a simple and general transfer learning framework that effectively leverages pre-trained language models to jointly learn DST and dialogue response generation. The $Lev$ is proposed for reducing the DST complexity and improving inference efficiency. In addition, two pre-trained Seq2Seq language models: T5~\cite{raffel2019exploring} and BART~\cite{lewis2019bart} are incorporated in our framework. Experimental results on MultiWOZ shows that, by using \textit{MinTL}, our systems not only achieve new SOTA result on both dialogue state tracking and end-to-end response generation but also improves the inference efficiency. In future work, we plan to explore task-oriented dialogues domain-adaptive pre-training methods~\cite{wu2020tod,peng2020soloist} to enhance our language model backbones, and extend the framework for mixed chit-chat and task-oriented dialogue agents~\cite{madotto2020adapter}.

\section{Acknowledgements}

This work has been partially funded by MRP/055/18 of the Innovation Technology Commission, The Hong Kong SAR Government. We would like to thanks Yichi Zhang for helpful discussion.

\bibliography{emnlp2020}
\bibliographystyle{acl_natbib}

\appendix
\clearpage

\section{KB States}
Table \ref{kb_state} shows KB states that are categorized by the number of matching entities and booking availability. $T_1$, $T_2$ are thresholds of the number of match entities. We define $T_1=1$ and $T_2=3$ for train domain, $T_1=5$ and $T_2=10$ for other domains.

\begin{table}[!ht]
\begin{tabular}{@{}lll@{}}
\toprule
\textbf{KB States} & \textbf{Entity Match} & \textbf{Book Availability} \\ \midrule
KB1               & -                     & -                          \\
KB2               & 0                     & -                          \\
KB3               & $\leq T_1$          & -                          \\
KB4               & $\leq T_2$         & -                          \\
KB5               & \textgreater \text{ } $T_2$      & -                          \\
KB6               & -                     & fail                       \\
KB7               & 0                     & fail                       \\
KB8               & $\leq T_1$          & fail                       \\
KB9               & $\leq T_2$         & fail                       \\
KB10               & \textgreater \text{ } $T_2$      & fail                       \\
KB11              & -                     & success                    \\
KB12              & 0                     & success                    \\
KB13              & $\leq T_1$          & success                    \\
KB14              & $\leq T_2$         & success                    \\
KB15              & \textgreater \text{ } $T_2$      & success                    \\ \bottomrule
\end{tabular}
\caption{KB states categorized by the number of matching entities and booking availability. $T_1$ and $T_2$ are thresholds. We define $T_1=1$ and $T_2=3$ for train domain, $T_1=5$ and $T_2=10$ for other domains.}
\label{kb_state}
\end{table}

\section{Hyper-parameters}
We report our training hyper-parameters on each task, which includes context window size \texttt{w}, learning rate \texttt{lr}, and learning rate decay rate \texttt{lr-decay}. We decay the learning rate when the performance in validation set does not improve. All of models are trained on Nvidia V100.
\begin{table}[!ht]
\begin{tabular}{@{}lllll@{}}
\toprule
\textbf{Task}                                                                                                             & \textbf{Model} & \textbf{w} & \textbf{lr} & \textbf{lr-decay} \\ \midrule
\multicolumn{1}{l|}{\multirow{3}{*}{\textbf{\begin{tabular}[c]{@{}l@{}}End-to-End\\ Response\\ Generation\end{tabular}}}} & T5-small       & 2          & 6e-4        & 0.8               \\
\multicolumn{1}{l|}{}                                                                                                     & T5-base        & 2          & 6e-4        & 0.8               \\
\multicolumn{1}{l|}{}                                                                                                     & BART-large     & 2          & 3e-5        & 0.8               \\ \midrule
\multicolumn{1}{l|}{\multirow{3}{*}{\textbf{DST}}}                                                                        & T5-small       & 3          & 6e-4        & 0.8               \\
\multicolumn{1}{l|}{}                                                                                                     & T5-base        & 3          & 6e-4        & 0.8               \\
\multicolumn{1}{l|}{}                                                                                                     & BART-large     & 3          & 1e-5        & 0.8               \\ \bottomrule
\end{tabular}
\caption{This table reports the hyper-parameters used by each model. $w$: window size, \texttt{lr}: learning rate, \texttt{lr-decay}: learning rate decay rate per epoch.}
\end{table}

\section{Generated Examples}
Table \ref{T5_example} - \ref{bart_example} shows the end-to-end response generation examples by \textit{MinTL} with T5-small, T5-base, and BART-large backbones.

\begin{table*}[!ht]
\resizebox{0.98\textwidth}{!}{
\begin{tabular}{@{}lll@{}}
\toprule
\textbf{MUL0113}                 & \textbf{}       & \textbf{Dialogue}                                                                                                                                                                                             \\ \midrule
\multirow{5}{*}{Turn 1} & User            & i am in need of an expensive indian restaurant for dinner .                                                                                                                                                   \\
                        & Dialogue State & {[}restaurant{]} food indian pricerange expensive                                                                                                                                                             \\
                        & Delex. Response  & \begin{tabular}[c]{@{}l@{}}there are {[}value\_choice{]} {[}value\_price{]} {[}value\_food{]} restaurants . \\ do you have a preference for area ?\end{tabular}                                               \\
                        & Lex. Response     & There are 14 expensive indian restaurants . Do you have a preference for area ?                                                                                                                               \\
                        & Gold Response   & there are 14 such restaurants . do you want to be in a specific part of town ?                                                                                                                                \\ \midrule
\multirow{5}{*}{Turn 2} & User            & i don't care what part of town it is in , i would like a phone number please .                                                                                                                                \\
                        & Dialogue State & {[}restaurant{]} food indian pricerange expensive                                                                                                                                                             \\
                        & Delex. Response   & \begin{tabular}[c]{@{}l@{}}{[}value\_name{]} is in the {[}value\_area{]} and is in the {[}value\_price{]} price range .\\ their phone number is {[}value\_phone{]} .\end{tabular}                             \\
                        & Lex. Response     & \begin{tabular}[c]{@{}l@{}}Curry Garden is in the centre and is in the expensive price range . \\ Their phone number is 01223302330 .\end{tabular}                                                            \\
                        & Gold response   & \begin{tabular}[c]{@{}l@{}}sure , i am sorry . i have curry garden their phone number is 01223302330 .\\ would you like me to reserve a table for you ?\end{tabular}                                          \\ \midrule
\multirow{5}{*}{Turn 3} & User            & not at this time . i am looking for a 3-star hotel , though . no guest house .                                                                                                                                \\
                        & Dialogue State & {[}restaurant{]} food indian pricerange expensive {[}hotel{]} stars 3 type hotel                                                                                                                              \\
                        & Delex. Response   & \begin{tabular}[c]{@{}l@{}}there are {[}value\_choice{]} {[}value\_type{]} that meet your criteria . \\ {[}value\_name{]} and {[}value\_name{]} . would you like me to book 1 of these for you ?\end{tabular} \\
                        & Lex. Response     & \begin{tabular}[c]{@{}l@{}}There are 2 hotel that meet your criteria . Gonville Hotel and the Lensfield Hotel .\\ Would you like me to book 1 of these for you ?\end{tabular}                                 \\
                        & Gold response   & \begin{tabular}[c]{@{}l@{}}i have found the gonville hotel . it is a 3 star hotel . would you like me to \\ book it for you ?\end{tabular}                                                                    \\ \midrule
\multirow{5}{*}{Turn 4} & User            & not yet . is the gonville hotel in the same price range as the curry garden ?                                                                                                                                 \\
                        & Dialogue State & {[}restaurant{]} food indian pricerange expensive {[}hotel{]} stars 3 type hotel                                                                                                                              \\
                        & Delex. Response   & yes , it is in the {[}value\_price{]} price range . would you like me to book it for you ?                                                                                                                    \\
                        & Lex. Response     & yes , it is in the expensive price range . would you like me to book it for you ?                                                                                                                             \\
                        & Gold response   & yes , it is an expensive hotel .                                                                                                                                                                              \\ \midrule
\multirow{5}{*}{Turn 5} & User            & can you book a room for 2 people for 2 nights starting on sunday ?                                                                                                                                            \\
                        & Dialogue State & \begin{tabular}[c]{@{}l@{}}{[}restaurant{]} food indian pricerange expensive {[}hotel{]} stars 3 type hotel \\ people 2 day sunday stay 2\end{tabular}                                                        \\
                        & Delex. Response   & \begin{tabular}[c]{@{}l@{}}I am sorry , but the booking was unsuccessful . Would you like to try \\ another day or a shorter stay ?\end{tabular}                                                              \\
                        & Lex. Response     & \begin{tabular}[c]{@{}l@{}}I am sorry , but the booking was unsuccessful . Would you like to try\\ another day or a shorter stay ?\end{tabular}                                                               \\
                        & Gold response   & \begin{tabular}[c]{@{}l@{}}the booking was unsuccessful . would you like to try for another day or\\ a shorter stay ?\end{tabular}                                                                            \\ \midrule
\multirow{5}{*}{Turn 6} & User            & \begin{tabular}[c]{@{}l@{}}could you try it for 1 night ? if that works , please book it and get the \\ reference number .\end{tabular}                                                                       \\
                        & Dialogue State & \begin{tabular}[c]{@{}l@{}}{[}restaurant{]} food indian pricerange expensive {[}hotel{]} stars 3 type hotel \\ people 2 day sunday stay 1\end{tabular}                                                        \\
                        & Delex. Response   & booking was successful . reference number is  : {[}value\_reference{]} .                                                                                                                                      \\
                        & Lex. Response     & Booking was successful . Reference number is : 5VRJSMSE .                                                                                                                                                     \\
                        & Gold response   & \begin{tabular}[c]{@{}l@{}}you are all set for 1 night ( sunday ) , reference number is 6wvv053q . \\ would you like to try another location for monday night ?\end{tabular}                                  \\ \midrule
\multirow{5}{*}{Turn 7} & User            & no . that is all i need , thank you .                                                                                                                                                                         \\
                        & Dialogue State & \begin{tabular}[c]{@{}l@{}}{[}restaurant{]} food indian pricerange expensive {[}hotel{]} stars 3 type hotel\\  people 2 day sunday stay 1\end{tabular}                                                        \\
                        & Delex. Response   & You are welcome . Have a great day .                                                                                                                                                                          \\
                        & Lex. Response     & You are welcome . Have a great day .                                                                                                                                                                          \\
                        & Gold response   & thank you , and goodbye .                                                                                                                                                                                     \\ \bottomrule
\end{tabular}
}
\caption{\textit{MinTL} (T5-small) end-to-end response generation on the test set of MultiWOZ. We show the dialogue states, delexicalized responses, lexicalized responses generated by our model.}
\label{T5_example}
\end{table*}

\begin{table*}[!ht]
\resizebox{0.98\textwidth}{!}{
\begin{tabular}{@{}lll@{}}
\toprule
\textbf{PMUL2000}       & \textbf{}       & \textbf{Dialogue}                                                                                                                                                                                                 \\ \midrule
\multirow{5}{*}{Turn 1} & User            & i am looking for information on a hotel called hobsons house .                                                                                                                                                    \\
                        & Dialogue State & {[}hotel{]} name hobsons house                                                                                                                                                                                    \\
                        & Delex. Response   & \begin{tabular}[c]{@{}l@{}}{[}value\_name{]} is a {[}value\_type{]} in the {[}value\_area{]} . it is {[}value\_price{]} and has\\ a {[}value\_stars{]} star rating . would you like to book a room ?\end{tabular} \\
                        & Lex. Response     & \begin{tabular}[c]{@{}l@{}}Hobsons House is a guest house in the west . It is moderate and has a \\ 3 star rating . Would you like to book a room ?\end{tabular}                                                  \\
                        & Gold response   & \begin{tabular}[c]{@{}l@{}}hobsons house is a guest house found in the west side of town . \\ it is in the moderate price range and is rated 3 stars .\end{tabular}                                               \\ \midrule
\multirow{5}{*}{Turn 2} & User            & do they have free parking ?                                                                                                                                                                                       \\
                        & Dialogue State & {[}hotel{]} name hobsons house                                                                                                                                                                                    \\
                        & Delex. Response   & yes , they do have free parking .                                                                                                                                                                                 \\
                        & Lex. Response     & Yes , they do have free parking .                                                                                                                                                                                 \\
                        & Gold response   & \begin{tabular}[c]{@{}l@{}}yes hobsons house has both free internet and parking . \\ would you like to book a room ?\end{tabular}                                                                                 \\ \midrule
\multirow{5}{*}{Turn 3} & User            & can i please get their postcode ?                                                                                                                                                                                 \\
                        & Dialogue State & {[}hotel{]} name hobsons house                                                                                                                                                                                    \\
                        & Delex. Response   & the postcode is {[}value\_postcode{]} . is there anything else i can help you with today?                                                                                                                         \\
                        & Lex. Response     & The postcode is CB39LH . Is there anything else i can help you with today ?                                                                                                                                       \\
                        & Gold response   & yes , of course ! it is cb39lh .                                                                                                                                                                                  \\ \midrule
\multirow{5}{*}{Turn 4} & User            & i also need a train that should depart from kings lynn and should go to cambridge .                                                                                                                               \\
                        & Dialogue State & {[}hotel{]} name hobsons house {[}train{]} destination cambridge departure kings lynn                                                                                                                             \\
                        & Delex. Response   & \begin{tabular}[c]{@{}l@{}}there are {[}value\_choice{]} trains that leave from {[}value\_departure{]} to \\ {[}value\_destination{]} . what day would you like to travel ?\end{tabular}                          \\
                        & Lex. Response     & \begin{tabular}[c]{@{}l@{}}There are 133 trains that leave from kings lynn to cambridge . \\ What day would you like to travel ?\end{tabular}                                                                     \\
                        & Gold response   & what day will you travel ?                                                                                                                                                                                        \\ \midrule
\multirow{5}{*}{Turn 5} & User            & i 'll be traveling on monday and need to arrive by 10 :00 .                                                                                                                                                       \\
                        & Dialogue State & \begin{tabular}[c]{@{}l@{}}{[}hotel{]} name hobsons house {[}train{]} destination cambridge departure kings lynn\\  day monday arrive 10 :00\end{tabular}                                                         \\
                        & Delex. Response   & \begin{tabular}[c]{@{}l@{}}the {[}value\_id{]} leaves at {[}value\_leave{]} and arrives at {[}value\_arrive{]} . \\ would you like me to book that for you ?\end{tabular}                                         \\
                        & Lex. Response     & \begin{tabular}[c]{@{}l@{}}The TR4943 leaves at 05:11 and arrives at 05:58 . Would you like\\ me to book that for you ?\end{tabular}                                                                              \\
                        & Gold response   & \begin{tabular}[c]{@{}l@{}}the tr0942 leaves kings lynn at 09 :11 and arrives in cambridge at 09 :58 . \\ would you like to book a reservation ?\end{tabular}                                                     \\ \midrule
\multirow{5}{*}{Turn 6} & User            & yes can i book it for 8 people ?                                                                                                                                                                                  \\
                        & Dialogue State & \begin{tabular}[c]{@{}l@{}}{[}hotel{]} name hobsons house {[}train{]} destination cambridge departure kings lynn \\ day monday arrive 10 :00 people 8\end{tabular}                                                \\
                        & Delex. Response   & \begin{tabular}[c]{@{}l@{}}booking was successful , the total fee is {[}value\_price{]} payable at the station . \\ reference number is  : {[}value\_reference{]} .\end{tabular}                                  \\
                        & Lex. Response     & \begin{tabular}[c]{@{}l@{}}Booking was successful , the total fee is 9.80 pounds payable at the station . \\ Reference number is : I4AFI8ET .\end{tabular}                                                        \\
                        & Gold response   & \begin{tabular}[c]{@{}l@{}}you are all set ! your reference \# is 2pffj67j . total cost is 78 .40 gbp , payable \\ at the station . can i help you with anything else today ?\end{tabular}                        \\ \midrule
\multirow{5}{*}{Turn 7} & User            & no , thank you . i have everything i need .                                                                                                                                                                       \\
                        & Dialogue State & \begin{tabular}[c]{@{}l@{}}{[}hotel{]} name hobsons house {[}train{]} destination cambridge departure kings lynn\\ day monday arrive 10 :00 people 8\end{tabular}                                                 \\
                        & Delex. Response   & you are welcome . have a great day .                                                                                                                                                                              \\
                        & Lex. Response     & You are welcome . Have a great day .                                                                                                                                                                              \\
                        & Gold response   & you have a wonderful day . goodbye .                                                                                                                                                                              \\ \bottomrule
\end{tabular}
}
\caption{\textit{MinTL} (T5-base) end-to-end response generation on the test set of MultiWOZ. We show the dialogue states, delexicalized responses, lexicalized responses generated by our model.}
\end{table*}

\begin{table*}[]
\resizebox{0.98\textwidth}{!}{
\begin{tabular}{@{}lll@{}}
\toprule
\textbf{MUL0570}        & \textbf{}       & \textbf{Dialogue}                                                                                                                                                         \\ \midrule
\multirow{5}{*}{Turn 1} & User            & hello , do you know of any points of interest in the centre of cambridge ?                                                                                                \\
                        & Dialogue state: & {[}attraction{]} area centre                                                                                                                                              \\
                        & Delexicalized   & \begin{tabular}[c]{@{}l@{}}there are {[}value\_choice{]} attractions in the {[}value\_area{]}  . what type of\\ attraction are you interested in ?\end{tabular}           \\
                        & lexicalized     & There are 44 attractions in the centre . What type of attraction are you interested in ?                                                                                  \\
                        & Gold response   & there are 44 attractions in the center of town . what types are you interested in ?                                                                                       \\ \midrule
\multirow{5}{*}{Turn 2} & User            & i am not sure . what types of attractions are there ?                                                                                                                     \\
                        & Dialogue state: & {[}attraction{]} area centre                                                                                                                                              \\
                        & Delexicalized   & there are {[}value\_type{]}  , {[}value\_type{]}  , and {[}value\_type{]}  .                                                                                              \\
                        & Lexicalized     & There are theatre , architecture , and nightclub .                                                                                                                        \\
                        & Gold response   & \begin{tabular}[c]{@{}l@{}}you could visit a church or college . there are also 8 museums that are free of \\ charge to enter .\end{tabular}                              \\ \midrule
\multirow{5}{*}{Turn 3} & User            & i would like to visit a college . may i have the entrance fee and postcard ?                                                                                              \\
                        & Dialogue state: & {[}attraction{]} area centre type college                                                                                                                                 \\
                        & Delexicalized   & \begin{tabular}[c]{@{}l@{}}there are {[}value\_choice{]} {[}value\_type{]} in the {[}value\_area{]}  . {[}value\_name{]} is \\ {[}value\_price{]} to enter .\end{tabular} \\
                        & Lexicalized     & There are 13 college in the centre . Christ 's College is free to enter .                                                                                                 \\
                        & Gold response   & there is christ 's college , which has a free entrance fee . their postcode is cb23bu .                                                                                   \\ \midrule
\multirow{5}{*}{Turn 4} & User            & \begin{tabular}[c]{@{}l@{}}can i have information for a train departing from ely ? i am looking to arrive in \\ cambridge by about 15 :15 on tuesday .\end{tabular}       \\
                        & Dialogue state: & \begin{tabular}[c]{@{}l@{}}{[}attraction{]} area centre type college {[}train{]} destination cambridge day tuesday \\ arrive 15 :15 departure ely\end{tabular}            \\
                        & Delexicalized   & there are {[}value\_choice{]} trains that fit your needs . what time would you like to leave ?                                                                            \\
                        & Lexicalized     & There are 5 trains that fit your needs . What time would you like to leave ?                                                                                              \\
                        & Gold response   & tr3412 will arrive around 11 :52 on tuesday if you are interested in that .                                                                                               \\ \midrule
\multirow{5}{*}{Turn 5} & User            & i 'll take it for just me what is the travel time , and the price please ?                                                                                                \\
                        & Dialogue state: & \begin{tabular}[c]{@{}l@{}}{[}attraction{]} area centre type college {[}train{]} destination cambridge day tuesday \\ arrive 15 :15 departure ely\end{tabular}            \\
                        & Delexicalized   & the travel time is {[}value\_time{]} and the price is {[}value\_price{]}  .                                                                                               \\
                        & Lexicalized     & The travel time is 17 minutes and the price is 4.40 pounds .                                                                                                              \\
                        & Gold response   & the travel time is 17 minutes and the price is 4 .40 gbp .                                                                                                                \\ \midrule
\multirow{5}{*}{Turn 6} & User            & great thank you i think i have all the information i need                                                                                                                 \\
                        & Dialogue state: & \begin{tabular}[c]{@{}l@{}}{[}attraction{]} area centre type college {[}train{]} destination cambridge day tuesday \\ arrive 15 :15 departure ely\end{tabular}            \\
                        & Delexicalized   & Thank you for using our service today !                                                                                                                                   \\
                        & Lexicalized     & Thank you for using our service today !                                                                                                                                   \\
                        & Gold response   & would you like me to book that train for you ?                                                                                                                            \\ \midrule
\multirow{5}{*}{Turn 7} & User            & no thank you . appreciate the help !                                                                                                                                      \\
                        & Dialogue state: & \begin{tabular}[c]{@{}l@{}}{[}attraction{]} area centre type college {[}train{]} destination cambridge day tuesday \\ arrive 15 :15 departure ely\end{tabular}            \\
                        & Delexicalized   & thank you for using our service . have a great day . goodbye .                                                                                                            \\
                        & Lexicalized     & thank you for using our service . have a great day . goodbye .                                                                                                            \\
                        & Gold response   & you are welcome . have a good day !                                                                                                                                       \\ \bottomrule
\end{tabular}
}
\caption{\textit{MinTL} (BART-large) end-to-end response generation on the test set of MultiWOZ. We show the dialogue states, delexicalized responses, lexicalized responses generated by our model.}
\label{bart_example}
\end{table*}

\end{document}